\documentclass[10pt,twocolumn,letterpaper]{article}

\usepackage[pagenumbers]{cvpr} 

\usepackage[dvipsnames]{xcolor}

\definecolor{cvprblue}{rgb}{0.21,0.49,0.74}
\usepackage[pagebackref,breaklinks,colorlinks,citecolor=cvprblue]{hyperref}

\title{DiffSLVA: Harnessing Diffusion Models for Sign Language Video Anonymization}

\author{Zhaoyang Xia\\
Rutgers University\\
110 Frelinghuysen Road\\
Piscataway, NJ 08854\\
{\tt\small zx149@rutgers.edu}
\and
Carol Neidle\\
Boston University\\
621 Commonwealth Ave.\\
Boston, MA 02215\\
{\tt\small carol@bu.edu}
\and
Dimitris N. Metaxas\\
Rutgers University\\
110 Frelinghuysen Road\\
Piscataway, NJ 08854\\
{\tt\small dnm@cs.rutgers.edu}
}

\begin{document}
\maketitle

\begin{abstract}
Since American Sign Language (ASL) has no standard written form, Deaf signers frequently share videos in order to communicate in their native language. However, since both hands and face convey critical linguistic information in signed languages, sign language videos cannot preserve signer privacy. While signers have expressed interest, for a variety of applications, in sign language video anonymization that would effectively preserve linguistic content, attempts to develop such technology have had limited success, given the complexity of hand movements and facial expressions. Existing approaches rely predominantly on precise pose estimations of the signer in video footage and often require sign language video datasets for training. These requirements prevent them from processing videos 'in the wild,' in part because of the limited diversity present in current sign language video datasets. To address these limitations, our research introduces DiffSLVA, a novel methodology that utilizes pre-trained large-scale diffusion models for zero-shot text-guided sign language video anonymization. We incorporate ControlNet, which leverages low-level image features such as HED (Holistically-Nested Edge Detection) edges, to circumvent the need for pose estimation. Additionally, we develop a specialized module dedicated to capturing facial expressions, which are critical  for conveying essential linguistic information in signed languages. We then combine the above methods to achieve anonymization that better preserves the essential linguistic content of the original signer. This innovative methodology makes possible, for the first time, sign language video anonymization that could be used for real-world applications, which would offer significant benefits to the Deaf and Hard-of-Hearing communities. We demonstrate the effectiveness of our approach with a series of signer anonymization experiments. 
\end{abstract}    
\section{Introduction}
\label{sec:intro}
\begin{figure*}[htbp]
\begin{center}
\includegraphics[scale=0.37]{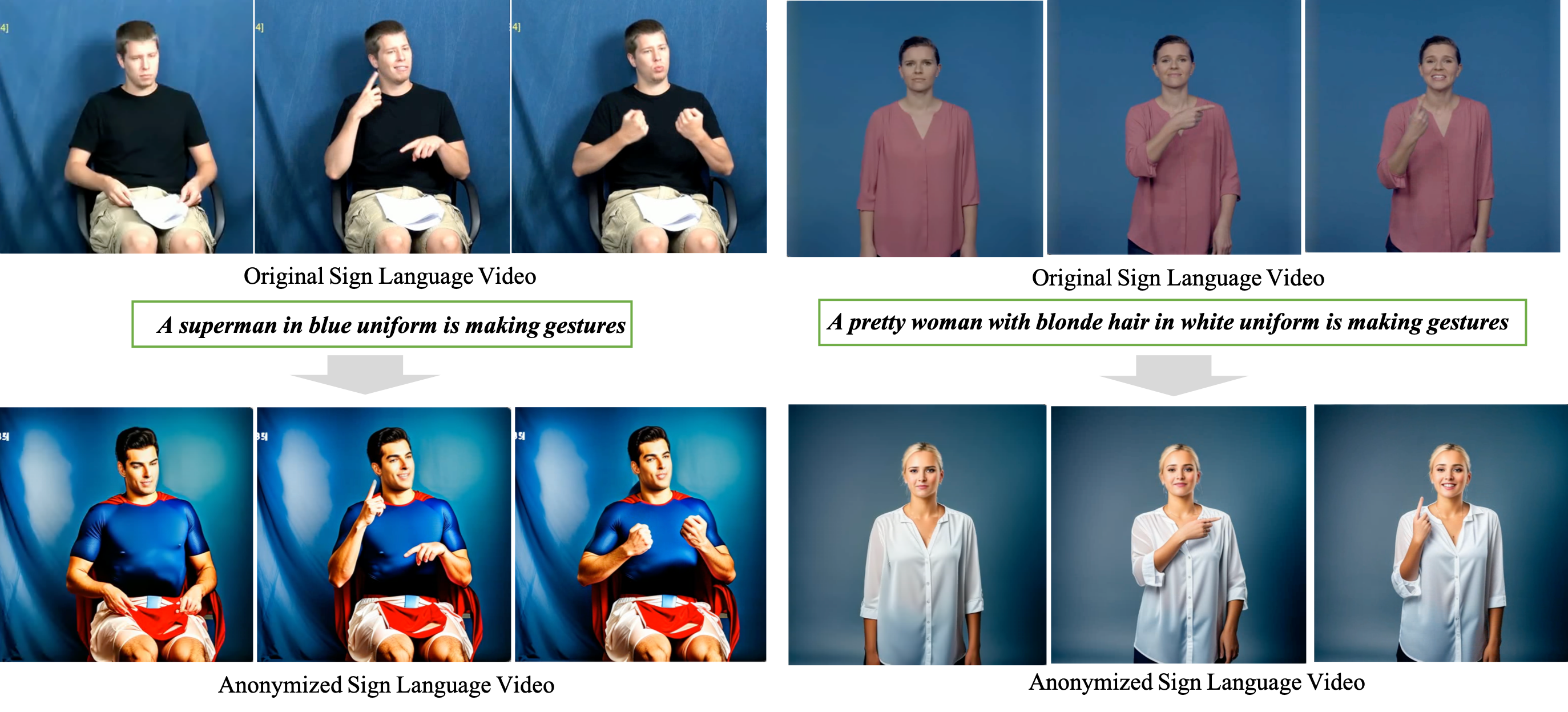} 
\caption{\textbf{Text-guided Sign Language Video Anonymization.} We introduce DiffSLVA, an innovative approach that leverages the capabilities of diffusion models to achieve text-guided sign language video anonymization. This method is capable of anonymizing sign language videos with a single text prompt, effectively masking the identity of the original signer while preserving the linguistic content and nuances.}
\label{fig.intro}
\end{center}
\end{figure*}

American Sign Language (ASL), the predominant form of communication used by the Deaf Community in the United States and parts of Canada, is a full-fledged natural language. It employs manual signs in parallel with non-manual elements, including facial expressions and movements of the head and upper body, to convey linguistic information. The non-manual elements are crucial for conveying many types of lexical and adverbial information, as well as for marking syntactic structures (e.g., negation, topics, question status, and clause types \cite{baker1985facial,kacorri2016continuous,neidle2000syntax,coulter1979american,valli2000linguistics}). Consequently, in video communications, e.g., on the Web, involving sensitive subjects such as medical, legal, or controversial matters, obscuring the face for purposes of anonymity would result in significant loss of essential linguistic information.

Despite the fact that a number of writing systems have been developed for ASL \cite{arnold2009proposal}, the language has no standard written form. While ASL signers could choose to use written English in order to preserve privacy, that is frequently not their preference, as signers generally have greater ease and fluency in their native language, ASL, than in English.

A considerable number of Deaf signers have shown interest in a mechanism that would maintain the integrity of linguistic content in ASL videos while disguising the identity of the signer, as discussed in several recent studies \cite{10.1145/3441852.3471200}. There are  many potential applications of such a tool. For example, this could enable anonymous peer review for academic submissions in ASL. This could also ensure impartiality in various multimodal ASL-based applications, e.g., enabling production of neutral definitions for ASL dictionaries, not tied to the identity of the signer producing them. It could also enable maintenance of neutrality in interpretation scenarios. Additionally, such a tool could increase signers' willingness to contribute to video-based AI datasets \cite{bragg2020exploring}, which hold significant research value.

For these reasons, various approaches for preservation of privacy in ASL videos have been explored \cite{isard2020approaches}. However, the majority of these approaches suffer from limitations with respect to preservation of linguistic meaning, and they generally achieve only a limited degree of anonymity.  They also require accurate pose estimation, and some require substantial human labor.  Furthermore, the effectiveness of many existing anonymization tools is limited to experimental settings, displaying sub-optimal performance with out-of-domain videos. These limitations significantly reduce the potential for practical applications of such technologies.

To overcome the limitations of existing anonymization tools, we introduce DiffSLVA, a novel anonymization approach leveraging large-scale pre-trained diffusion models, notably Stable Diffusion \cite{rombach2022high}. DiffSLVA is designed to tackle text-guided sign language anonymization. Through a text prompt, it generates a new video in which the original linguistic meaning is retained, but the identity of the signer is altered. See Figure \ref{fig.intro} for a demonstration of the method. Unlike traditional methods that require skeleton extraction, our approach utilizes the Stable Diffusion model enhanced with ControlNet \cite{zhang2023adding} to process language videos with Holistically-Nested Edge (HED) \cite{xie2015holistically}, which can much more easily and robustly process videos in the wild. To adapt the image-based Stable Diffusion for video, we follow \cite{yang2023rerender} but modify its architecture. We replace the self-attention layer in U-Net with a cross-frame attention layer and implement an optical-flow guided latent fusion for consistent frame generation. Additionally, to capture fine-grained facial expressions, we have developed a specialized facial generation module utilizing a state-of-the-art image animation model \cite{zhao2022thin}. The outcomes are integrated via a face segmentation technique \cite{yu2018bisenet}. Our  results show substantial promise for anonymization applications in the wild, which would be invaluable for the Deaf and Hard-of-Hearing communities. 


Our work makes several key contributions to the field of sign language video anonymization:
\begin{enumerate}
\item We propose zero-shot text-guided sign language anonymization: We are the first to address the challenge of zero-shot sign language video anonymization. Our method does not require sign language video data for training. The anonymized videos are based on computer-generated humans, transforming the original signer's appearance to that of a computer-generated individual.
\item We have developed a specialized module dedicated to improving facial expression transformation. Our ablation studies show that this significantly enhances the preservation of linguistic meaning.
\item Our approach relies solely on low-level image features, such as edges, enhancing the potential for practical applications, which is a significant achievement.
\item Our anonymization can accommodate a diverse range of target humans. The anonymized signers can have any ethnic identity, gender, clothing, or facial style, a feature many ASL signers want; this simply requires changing the text input.
\end{enumerate}

\section{Related Work}
\label{sec:related_work}
\subsection{Video Editing with Diffusion Models}

Diffusion models \cite{ho2020denoising} have demonstrated exceptional performance in the field of generative AI. Once such models are trained on large-scale datasets (e.g., LAION \cite{schuhmann2022laion}), text-guided latent diffusion models \cite{rombach2022high} (e.g., Stable Diffusion) are capable of producing diverse and high-quality images from a single text prompt. Additionally, ControlNet \cite{zhang2023adding} presents a novel enhancement. It fine-tunes an additional input pathway for pre-trained latent diffusion models, enabling them to process various modalities, including edges, poses, and depth maps. This innovation significantly augments the spatial control capabilities of text-guided models.

Image-based diffusion models can also be used for video generation or editing. There have been efforts to modify image-based diffusion models for consistent generation or editing across frames. Tune-A-Video \cite{wu2023tune} inflates a pre-trained image diffusion model, modified with pseudo 3D convolution and cross-frame attention and then fine-tuned on a given video sequence. During the inference stage, with the DDIM inversion noises \cite{song2020denoising} as the starting point, the fine-tuned model is able to generate videos with similar motions but varied appearance. Edit-A-Video \cite{shin2023edit}, Video-P2P \cite{liu2023video}, and vid2vid-zero \cite{wang2023zero} utilize Null-Text Inversion \cite{mokady2023null} for improved reconstruction of video frames, which provides better editing results. Fine-tuning or optimization based on one or more input video sequences is required by these methods. Moreover, the detailed motion in the video cannot be captured properly without having a negative impact on the editing abilities. Therefore, they are not suitable for the sign language video anonymization task.

Other methods utilize the cross-frame attention mechanism or latent fusion to achieve the video editing or generation ability of image-based diffusion models. Text2Video-Zero \cite{khachatryan2023text2video} modifies the latent codes and attention layer. FateZero \cite{qi2023fatezero} blends the attention features based on the editing masks detected by Prompt-to-Prompt \cite{hertz2022prompt}. Pix2Video \cite{ceylan2023pix2video} aligns the latent features between frames for better consistency. Rerender-A-Video \cite{yang2023rerender} utilizes a cross-frame attention mechanism and cross-frame latent fusion to improve the consistency of style, texture, and details. It can also be used with ControlNet for spatial guidance. However, these methods cannot accurately translate facial expressions from the original videos. Therefore, they lose a significant amount of the linguistic meaning from the original video. Our approach is based on Rerender-A-Video \cite{yang2023rerender} method without the post video processing, to best capture manual signs. To overcome the loss of linguistically important non-manual information, we designed a specialized facial expression translation module \cite{zhao2022thin}, which we combine with the rest of the anonymized body
using a face parser model \cite{yu2018bisenet}.

\subsection{Sign Language Video Anonymization}

In the realm of privacy preservation in ASL video communication, various strategies have been investigated \cite{isard2020approaches}. Early approaches used graphical filters, such as a tiger-shaped filter \cite{bragg2020exploring}, to disguise the face during signing. However, these filters often lead to a loss of critical facial expressions, thereby hindering comprehension. Alternatives like blocking parts of the face \cite{bleicken-etal-2016-using} also result in significant information loss. Approaches involving re-enacting signed messages with actors \cite{isard2020approaches} or using virtual humans for anonymous sign language messaging \cite{heloir2016toward,efthimiou2015user} are labor-intensive, challenging, and time-consuming.

Some approaches to avatar generation for sign language, such as \cite{bragg2019sign}, have used cartoon-like characters to replace signers. Cartoonized Anonymization \cite{tze2022cartoonized} proposes the use of pose estimation models \cite{li2018crowdpose,xiu2018poseflow,lugaresi2019mediapipe} to automatically enable the avatars to sign. Yet, these methods often lead to unrealistic results \cite{kipp2011assessing}.

Deep-learning approaches, such as the AnonySign project \cite{saunders2021anonysign} or Neural Sign 
Reenactor \cite{tze2022neural}, leverage GAN-based methods for photo-realistic sign language anonymization using skeleton keypoints for accurate image generation. The results are encouraging. However, they require accurate skeleton keypoints and face landmarks. In sign language videos, the rapid movements of the hands can lead to blurring in the video frames. Occlusions of the face by the hands also occur frequently. The performance of existing human pose estimation models is often inadequate when applied to sign language videos, which leads to errors in the anonymized video. 

Recent work \cite{10.1145/3441852.3471200} applies the facial expression transfer method of \cite{siarohin2019first} for sign language anonymization. This method involves replacing the signer's face in the video with another individual's face, while transferring the facial expressions to the new face. As a result, this approach successfully preserves the linguistic meanings conveyed by facial expressions and alters the identity of the signer in the video. However, in \cite{10.1145/3441852.3471200} the extent of the anonymization is not complete, since only the face is replaced, while the arms, torso, and hands remain the same as in the original video. 
Another method \cite{xia2022sign} uses an unsupervised image animation method \cite{siarohin2021motion} with a high-resolution decoder and loss designed for the face and hands to transform the identity of a signer to that of another signer from the training videos. The results are promising. However, this method can  work well only in the training data domain and is hard to adapt to sign language videos in the wild. 

To address the above limitations, we propose DiffSLVA, a method that is based on the modification of large-scale diffuson models and ControlNet for consistent high-fidelity video generation, which can be used to achieve effective sign language video anonymization in the wild. Our approach is a text-guided sign language video anonymization, as shown in Figure \ref{fig.intro}. We use large-scale diffusion models, which do not rely on the use of sign language video data for training and can perform zero-shot sign language video anonymization in the wild. With the help of ControlNet, we use low-level features instead of accurate skeleton data as signal for generation guidance so that the results are not adversely affected by inaccurate skeleton estimations. To further improve the facial expression translation, we designed a specialized model for facial expression enhancement and combine it with the model that anonymizes the rest of the body
using a face parser model. Our method can anonymize sign language videos based on a single text prompt. The anonymized video is based only on a wide range of computer-generated humans. Our successful anonymization results in the wild show great promise for use by the Deaf community.


 
\section{Methodology}
\label{sec:methodology}

\begin{figure*}[htbp]
\begin{center}
\includegraphics[width=1\linewidth]{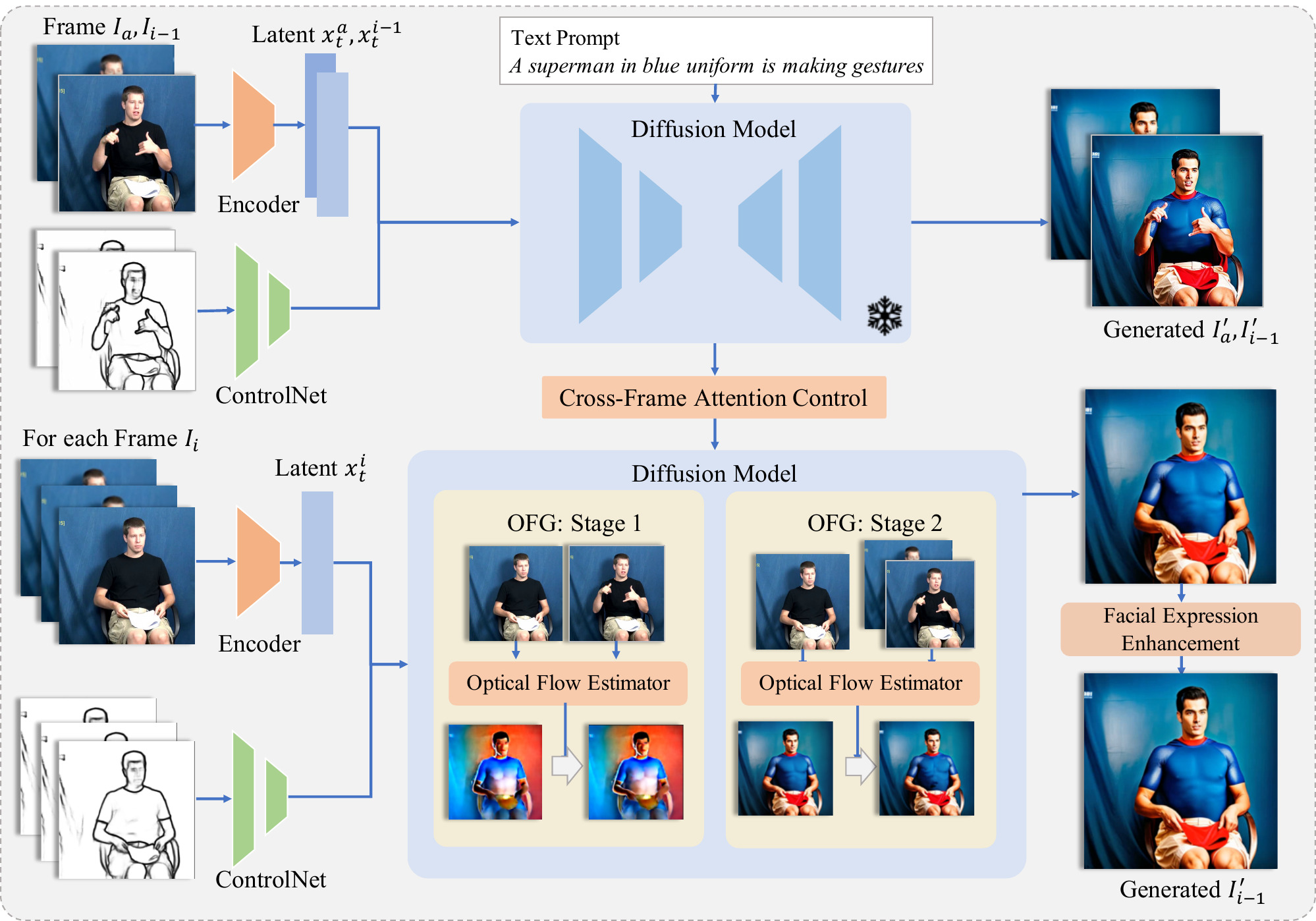} 
\caption{\textbf{Method Overview.} The original frames $\{I_i\},i=1,...,N$ in the sign language video are encoded and sampled as noisy latent features $\{x^{i}\}_t,i=1,...,N$. An anchor frame $I_a$ and its Holistically-Nested Edge are used to generate the $I'_a$ with ControlNet, which will constrain the global style consistency. For each frame $I_i$, the previous generated frame $I'_{i-1}$ and the anchor generated frame $I'_a$ provide cross-frame attention control during the generation process of $I'_i$. A two-stage optical flow guided latent fusion is applied. A specialized facial expression enhancement module is used to update $I'_i$ for the final result.}
\label{fig.flowchart}
\end{center}
\end{figure*}

In this section, we introduce our method for zero-shot text-guided sign language video anonymization. The process is structured as follows: Given a sign language video with $N$ frames $\{I_i\}_{i=0}^N$, we employ a pre-trained latent diffusion model augmented with ControlNet to execute the anonymization. A text prompt $c_p$ serves as the guidance for the desired anonymization identity or style. Our goal is to generate an altered sign language video sequence, represented by $\{I'_i\}_{i=0}^N$, which conceals the identity of the original signer while preserving the linguistic content. 

In section \ref{sec:ldm}, we introduce the text-guided latent diffusion models and the ControlNet, which serve as the foundation for text-guided image generation. Section \ref{sec:video} details the methods for adapting the text-to-image method for consistent video editing. To ensure the preservation of linguistic meaning through accurate facial expression translation, we introduce a specialized facial enhancement module in Section \ref{sec:face}. Figure \ref{fig.flowchart} shows an overview of our method.
\subsection{Latent Diffusion Models}
\label{sec:ldm}
Latent diffusion models are diffusion models operating in the latent space for faster image generation. One major feature of the approach is that it uses an autoencoder, U-Net, and a text encoder.
One difference with respect to the standard forward and denoising process is that the input image $I$ is first input to an encoder $\varepsilon$ to obtain its latent features $x_0 = \varepsilon (I)$. The following diffusion forward process adds noise to the latent features
\begin{equation}
  q(x_t|x_{t-1}) = \mathcal{N}(x_t;\sqrt{\alpha_t}x_{t-1},(1-\alpha_t)\mathbf{I}),
  \label{eq:eq1}
\end{equation}
where $t=1,...,T$ is the time step indicating the level of noises added;  $q(x_t|x_{t-1})$ is the conditional probability of $x_t$ given $x_{t-1}$; and $\alpha_t$ are hyperparameters that adjust the noise level across the time step $t$. Leveraging the property of Gaussian noise, we can also sample $x_t$ at any time step by the following equation:
\begin{equation}
  q(x_t|x_0) = \mathcal{N}(x_t;\sqrt{\bar{\alpha}_t}x_0,(1-\bar{\alpha}_t)\mathbf{I}),
  \label{eq:eq1}
\end{equation}
where $\bar{\alpha}_t = \prod_{i=1}^t \alpha_i$.

In the diffusion backward process, a U-Net $\epsilon_{\theta}$ is trained to estimate the above added noise to recover $x_0$ from $x_T$. For the conditional diffusion model, $\epsilon_{\theta}$ takes the conditional information $c_p$ as input to guide the generation process. After $\epsilon_{\theta}$ has been trained, the $x_{t-1}$ can be sampled by strategies such as DDIM sampling \cite{song2020denoising}:
\begin{equation}
  x_{t-1} = \sqrt{\bar{\alpha}_{t-1}}\hat{x}_0+\sqrt{1-{\bar\alpha}_{t-1}}\epsilon_{\theta}(x_t,t,c_p),
  \label{eq:eq1}
\end{equation}
where $\epsilon_{\theta}(x_t,t,c_p)$ is the predicted noise at time step $t$.
For the DDIM sampler, we can have an estimation of the final clear output $\hat{x}_0$ at each time step $t$. $\hat{x}_0$ can also be represented as the following equation:
\begin{equation}
  \hat{x}_0 = (x_t-\sqrt{1-{\bar\alpha}_{t}}\epsilon_{\theta}(x_t,t,c_p))/\sqrt{\bar\alpha_t},
  \label{eq:x0}
\end{equation}
During inference, for a Gaussion noise $x_T$, we can sample a clear latent $x_0$ with the DDIM Sampler and decode it to the generated image $I'=D(x_0)$

Our methodology also incorporates ControlNet, which is inspired by the Hyper Network concept. ControlNet introduces an additional signal to the text-guided latent diffusion models. This structure makes it possible for the text-guided diffusion model to take diverse inputs like edges, human poses, and segmentation maps for more spatial constraints. Consequently, with the incorporation of an additional input $c_n$, the predicted noise at each time step $t$ is represented as $\epsilon_{\theta}(x_t, t, c_p, c_n)$. This approach enhances the alignment of the final outputs with the spatial features specified by the input condition $c_n$.

\subsection{Consistent Video Generation}
\label{sec:video}
Although Stable Diffusion models exhibit outstanding performance in image generation, their direct application to videos is challenging. Directly applying Stable Diffusion to videos gives rise to significant frame inconsistency issues. To address this, we adapt text-to-image diffusion models for video editing tasks, drawing upon the framework established by \cite{yang2023rerender}. Our approach begins by encoding and sampling the original frames ${I_i}, i = 1, \ldots, N$, of the sign language video into noisy latents ${x^i}_t, i = 1, \ldots, N$, serving as starting points for the generation of anonymized video frames, following the method described in \cite{meng2021sdedit}. An anchor frame $I_a$ is selected from the sequence ${I_i}, i = 1, \ldots, N$. The corresponding latent feature $x^a_t$, along with the Holistically-Nested Edge, is processed through ControlNet to create the transformed anchor frame $I'a$, which constraints the global consistency in general. Empirically, we find that selecting the anchor frame from the middle of the video, where both hands of the signer are visible, yields optimal results. For each frame $I_i$, the previously generated frame $I'_{i-1}$ and the anchor frame $I'_a$ provide cross-frame attention control during the generation of $I'_i$, as detailed in Section \ref{sec:cfc}. A two-stage optical flow guided latent fusion, described in Section \ref{sec:tlf}, is applied during the generation process. Finally, a specialized facial expression enhancement module, outlined in Section \ref{sec:face}, is used to refine the results.

\subsubsection{Cross-Frame Attention Consistency}
\label{sec:cfc}
In the Stable Diffusion model, there are two kinds of attention mechanisms used in the U-Net. The cross-attention retrieves the information from the text embedding. The self-attention helps define the layout and style of the generated images. In order to achieve consistent generation across frames in the sign language video sequence, the self-attention layers are replaced with cross-frame attention layers. The self-attention layer of the U-Net used in Stable Diffusion is represented as follows:
\begin{equation}
  Q=W^Q v_i, K=W^K v_i, V=W^V v_i,
  \label{eq:eq1}
\end{equation}
where $v_i$ is the latent features input to the self-attention layer when generating $I'_i$. $W^Q$, $W^K$, and $W^V$ are the weights for project $v_i$ to the query, key, and value in the attention mechanism, respectively.
The attention map $SA$ is calculated as following:
\begin{equation}
  SA(Q,K,V) = \text{Softmax}(\frac{QK^T}{\sqrt{d}})  V
  \label{eq:eq1}
\end{equation}
In order to obtain consistent generation across frames, we replace the $K$ and $V$ with $K_{a,i-1}$ and $V_{a,i-1}$, which are the combination of keys and values when generating the selected anchor frame $I_a$ and previous frame $I_{i-1}$. The cross-frame attention layer is represented as follows:
\begin{align}
  K_{a,i-1}&=W^K [v_a;v_{i-1}],\quad Q=W^Q v_i\nonumber\\
  V_{a,i-1}&=W^V [v_a;v_{i-1}],
  \label{eq:eq1}
\end{align}
where $v_a$, $v_{i-1}$ are the latent features obtained when generating frame $I'_a$ and $I'_{i-1}$. The cross attention map $CA$ is calculated as following:
\begin{equation}
  CA(Q,K_{a,i-1},V_{a,i-1}) = \text{Softmax}(\frac{QK_{a,i-1}^T}{\sqrt{d}})V_{a,i-1}
  \label{eq:eq1}
\end{equation}
The cross-frame attention mechanism is designed to foster consistency in image generation across frames by directing the current generation process to reference patches in both the generated anchor frame and the previous frame.

\subsubsection{Optical Flow Guided Cross-Frame Latent Fusion}
\label{sec:tlf}
Following \cite{yang2023rerender}, we utilize two-stage latent fusion guided by optical flow: OFG stage 1 and OFG stage 2.

OFG stage 1: In the early stage of the diffusion backward process, the optical flow $w^i_a$ and occlusion mask $M^i_a$ are estimated from $I_a$ to $I_i$ to wrap and fuse
the estimated latent of $I'_a$ and $I'_i$. This latent wrap and fusion is performed when the denoising step $t$ is large, to prevent distortion of the results. At time step $t$, the predicted $\hat{x}_0$ is updated by the following equation:
\begin{equation}
  \hat{x}^i_0 = M^i_a \hat{x}^i_0 + (1-M^i_a)w^i_a(\hat{x}^a_0),
  \label{eq:eq1}
\end{equation}
where $\hat{x}^i_0$ and $\hat{x}^a_0$ are the predicted clear outputs for $I'_i$ and $I'_a$ at denoising time step $t$, calculated by equation \ref{eq:x0}. 

OFG stage 2: At the second stage, the generated anchor frame $I'_a$ and previous generated frame $I'_{i-1}$ are used to further enhance consistency during the late stages of the diffusion backward process. The optical flow and occlusion mask are also estimated. We obtain a reference image $\bar{I'}_i$ by wrapping and fusing with the previous generated images:

\begin{equation}
  \bar{I'}_i = M^i_a(M^i_{i-1} \hat{I'}_i + (1-M^i_{i-1})w^i_{i-1}(I'_{i-1})) + (1-M^i_a)w^i_a I'_a,
  \label{eq:eq1}
\end{equation}
After obtaining this reference-estimated image $\bar{I'}_i$, we can update the sampling process for generating $I'_i$ using the following equation:
\begin{equation}
  x^i_{t-1} = M_i x^i_{t-1} + (1-M_i) \bar{x}^i_{t-1},
  \label{eq:eq1}
\end{equation}
where $M_i=M^i_a \cap M^i_{i-1}$, and $\bar{x}^i_{t-1}$ is the sampled $x_{t-1}$ from reference image $\bar{I'_i}$. We use the same strategy as the fidelity-oriented image encoding from \cite{yang2023rerender} for encoding the $\bar{I'_i}$ to avoid information loss when repeatedly encoding and decoding latents.

To maintain coherent color throughout the whole process, we also apply AdaIN\cite{huang2017arbitrary} to $\hat{x}^i_0$ with $\hat{x}^a_0$ at time step $t$ during the late stage of the diffusion backward process. This is used to mitigate the color draft problem with diffusion models.


\subsection{Facial Expression Enhancement}

\begin{figure}[htbp]
\begin{center}
\includegraphics[width=1\linewidth]{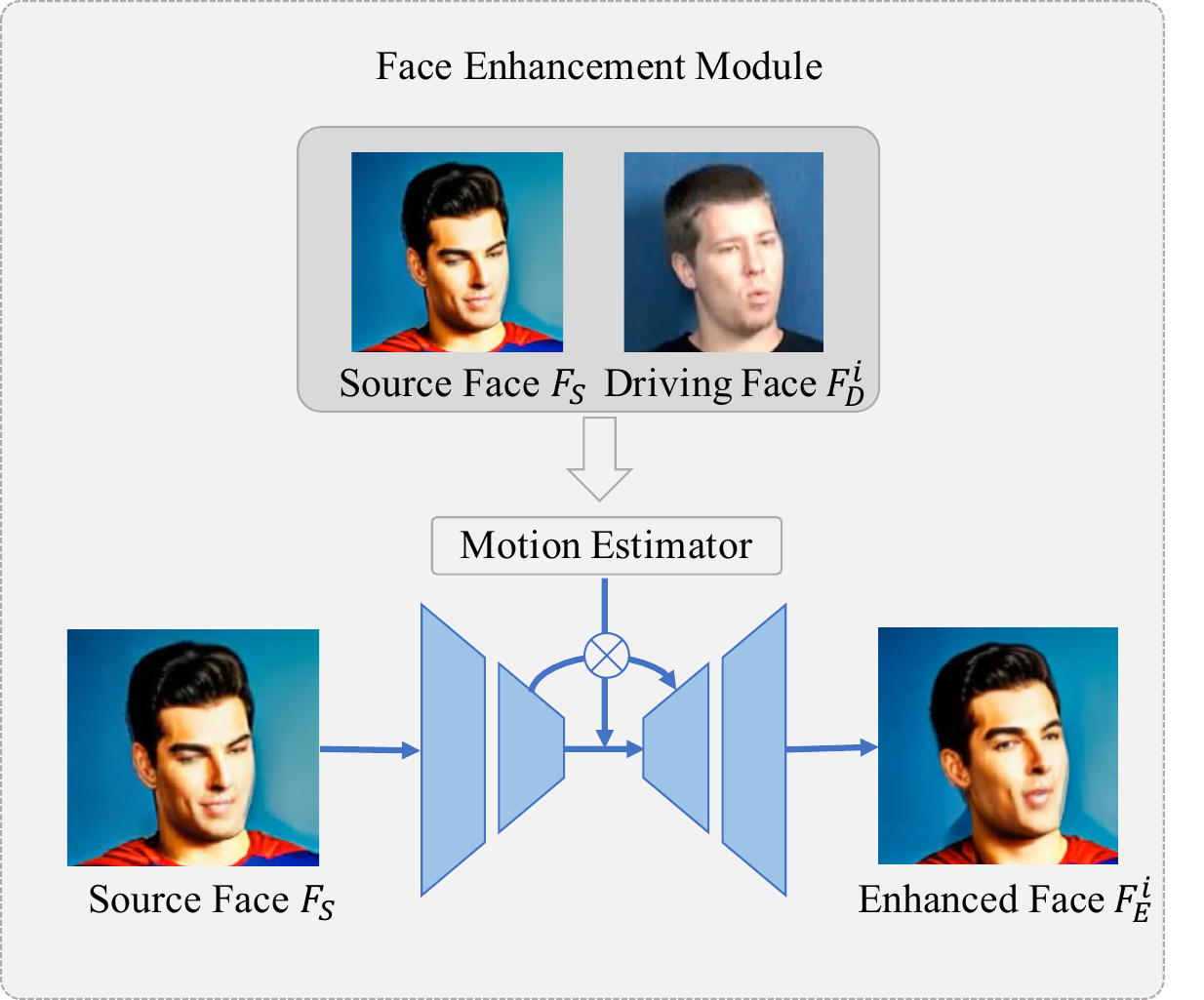} 
\caption{\textbf{Face Enhancement Module.} The motion estimator obtains a dense motion map and a multi-resolution occlusion map between the source face $F_s$ and the driving face. The output along with a U-Net is applied to generate the enhanced face $F^i_E$}
\label{fig.facechart}
\end{center}
\end{figure}

Facial expressions convey important linguistic meaning in signed languages. However, current methods cannot transfer meaningful facial expressions; see the ablation study discussed in Section \ref{sec:abl}. ControlNet and Stable Diffusion usually fail to produce faces with the same expressions as the original signer. To address this issue, we propose an additional module to enhance the face generation based on an image-animation model. See Figure \ref{fig.facechart} for an overview of this module.

When generating the first frame $I'_1$ , we crop the face of the results and use it as the source face $F_s$ for the image animation module from \cite{zhao2022thin}. The facial images in the original videos are also cropped and aligned to formalize the driving face set $[F^{i}_d],i=1...N$. A motion estimation module, which is pre-trained on Voxceleb\cite{nagrani2017voxceleb}, will estimate the dense motion $W_i$ and multi-resolution occlusion maps $M_i$ between the source face $F_s$ and the driving face set $[F^{i}_d],d=1...N$.

The obtained optical flow and occlusion maps are input to a U-Net to generate new face images that match the identity of the source face $F_s$ while having the same facial expression as $F^{i}_d$. The input image $F_s$ is processed through the encoder, and optical flow $W_i$ is applied to wrap the feature map at each level. This adjusted feature map is then combined with the occlusion mask $M^f_i$ that matches its resolution. Subsequently, it is merged into the decoder through a skip connection. After this, the feature map is input to the next upsampling layer. Finally, the enhanced face image $F^i_E$ is produced at the last layer.

A face parser model \cite{yu2018bisenet} is applied on $F^i_E$ to segment the face area and obtain a mask $M^f_i$. Then, the mask and enhanced face image are aligned with the face location in $I'_i$. Finally, $I'_i$ is updated by the following equation:

\begin{equation}
  I'_i = M^f_i F^i_E + (1-M^f_i) I'_i.
  \label{eq:eq1}
\end{equation}

\label{sec:face}

\section{Experiments and Results}
\label{sec:experiment}

\begin{figure*}[htbp]
\begin{center}
\includegraphics[width=0.8\linewidth]{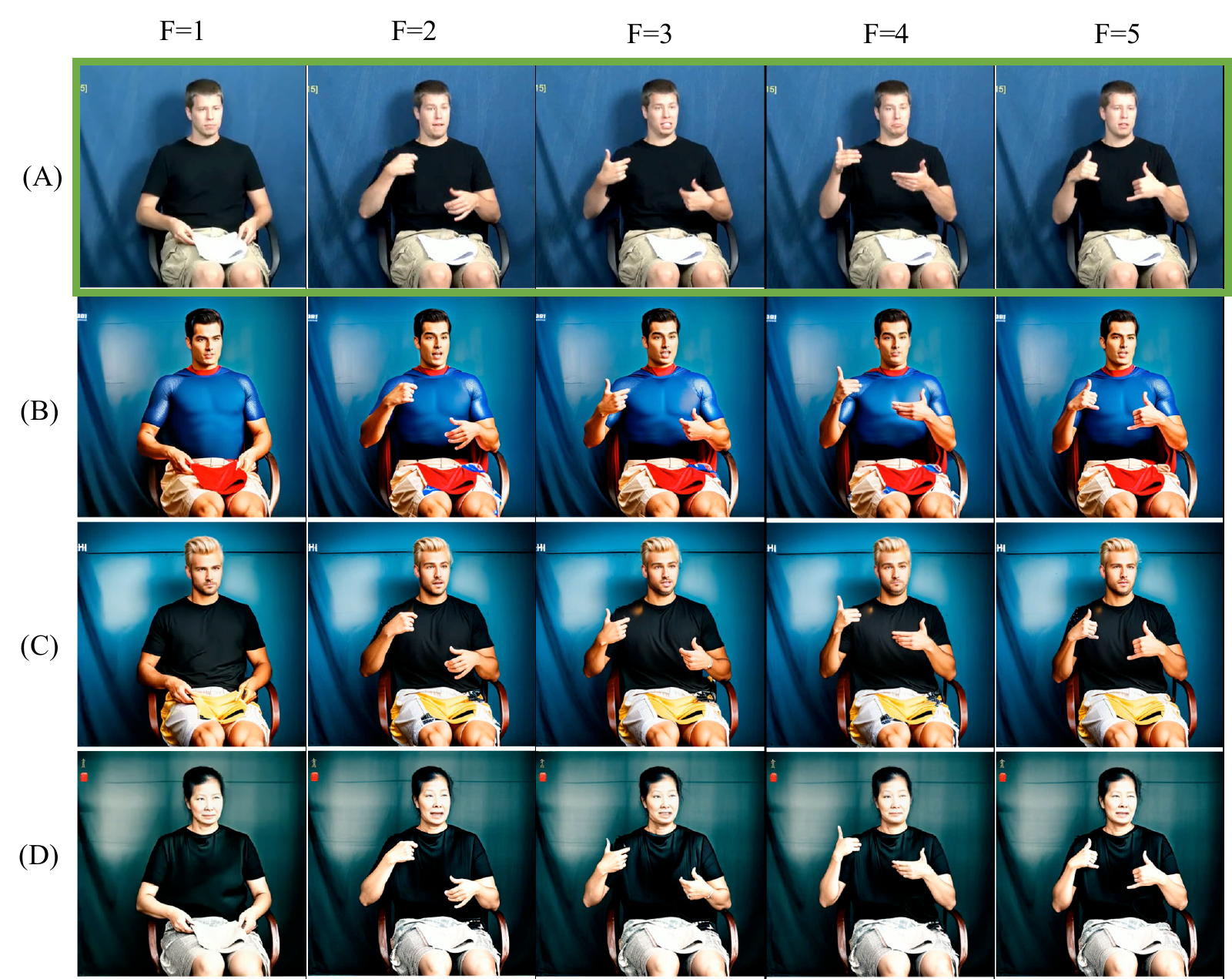} 
\caption{\textbf{Anonymization Result Examples.} Row (A) contains some frames from the original sign language video of the sentence meaning 'If friends play frisbee, I will join them in playing' (taken from the ASLLRP file Cory\_2013-6-27\_sc115, Utterance 22).
Rows (B)-(D) are the anonymized results using different prompts: (B) \textit{a Superman in blue uniform is making gestures} (C) \textit{a man in CG style, blond hair, is making gestures} (D) \textit{a woman in Chinese ink wash painting is making gestures.} }
\label{fig.res_1}
\end{center}
\end{figure*}
\begin{figure*}[htbp]
\begin{center}
\includegraphics[width=0.9\linewidth]{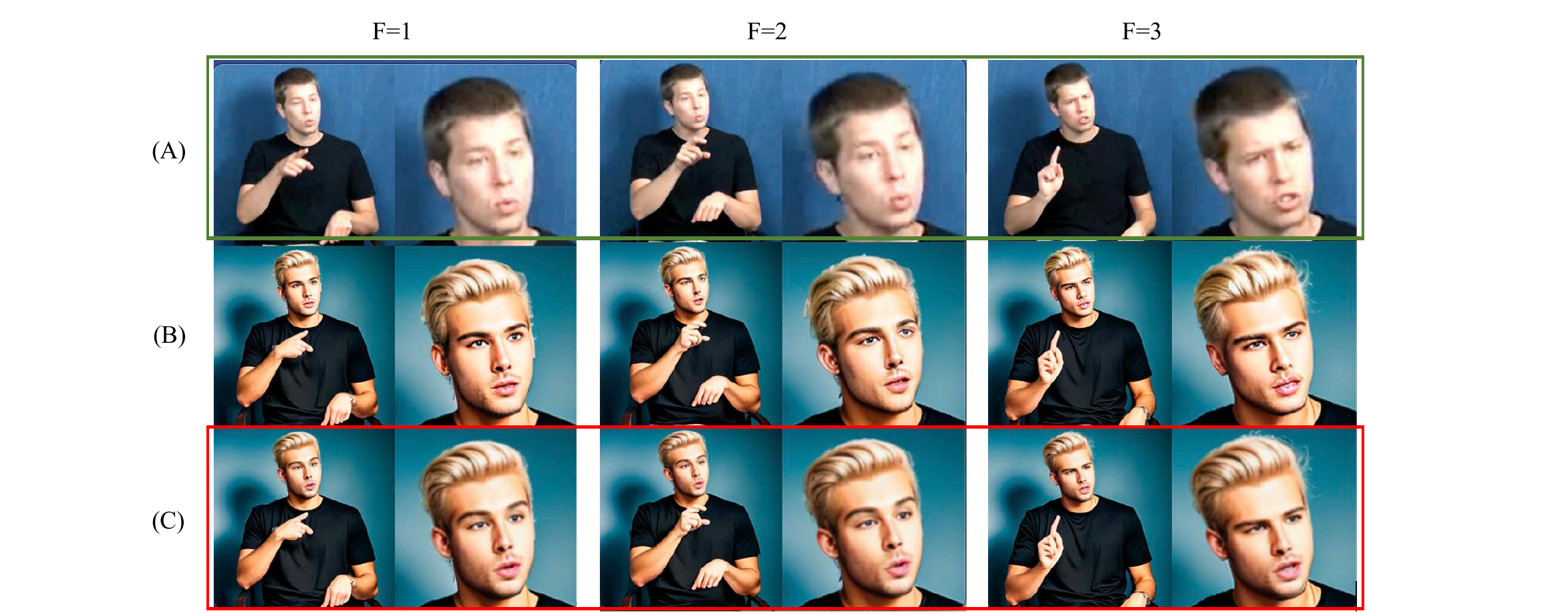} 
\caption{\textbf{Ablation Study of Facial Expression Enhancement}. The frames in Row (A) are from the original video of the ASL sentence meaning 'You work where?' (ASLLRP file Cory\_2013-6-27\_sc113, Utterance 28). Row (B) is the result without applying the facial enhancement module. Row (C) is the final result of our method.}
\label{fig.abl_1}
\end{center}
\end{figure*}

\subsection{Data Set}

We implemented our method on video datasets distributed through the American Sign Language Linguistic Research Project (ASLLRP): \href{https://dai.cs.rutgers.edu/dai/s/dai}{https://dai.cs.rutgers.edu/dai/s/dai} \cite{neidle2018new,neidle2022asl}. To assess the effectiveness of our anonymization technique, we selected signers of diverse genders and ages. Each test sample was limited to a maximum of 180 video frames. Example results are presented in Figure \ref{fig.res_1}. 

\subsection{Models}

Our experiments utilized Stable Diffusion models version 1.5 and other customized models. The ControlNet version 1.0 was employed, producing optimal results with HED as a conditional input. Optical flow estimation was performed using the model from \cite{xu2022gmflow}. 

\subsection{Qualitative Evaluation}

Overall, our method generates clear hand shapes with high fidelity to the original signer’s hand shapes and movement of the hands and arms.  Most of the generated facial expressions are good, and we are currently carrying out further refinements to fully preserve the subtleties of expressions that are critical to expression of linguistic information. The effectiveness of our combined method for transmission of linguistic content, complete disguise of identity, and production of natural-looking signing remains to be confirmed through  user studies, which we plan to carry out in the near future.  However, the initial results are quite encouraging.  As shown in Figure \ref{fig.res_1},  our methods, 
guided by text prompts, can anonymize original videos to computer-generated signers with different genders and identities:  
 With different text prompts, we can have various anonymized versions of the sign language videos, 
 from the CG  (Computer Graphics) style to ink washing painting. Some video examples can be viewed at \href{https://github.com/Jeffery9707/DiffSLVA}{https://github.com/Jeffery9707/DiffSLVA}. These results underscore the practical potential of our approach.

To our knowledge, this is the first instance of zero-shot sign language anonymization in real-world scenarios. Methods like Cartoonized Anonymization (CA) \cite{tze2022cartoonized} cannot generate photorealistic results and rely on skeleton estimation for accurate anonymization. Methods that can generate photorealistic results, such as AnonySign \cite{saunders2021anonysign}, SLA \cite{xia2022sign} and Neural Sign Reenactor (NSR) \cite{tze2022neural}, require training on sign language video datasets or accurate skeleton estimation. These methods are not accurate enough to be used in the wild.

\subsection{Ablation Study}
\label{sec:abl}

Our ablation study focused on the facial expression enhancement module. Results are illustrated in Figure \ref{fig.abl_1}. Using a separate module significantly improves the preservation of linguistic meaning; the example shown in this figure includes topic and wh-question marking. A video example is also available for viewing at \href{https://github.com/Jeffery9707/DiffSLVA}{https://github.com/Jeffery9707/DiffSLVA}. 

There is a notable challenge with the Stable Diffusion model, primarily in its ability to  generate varied facial expressions accurately for the sign language video anonymization task. Instead of producing diverse expressions, the model tends to replicate a uniform expression across different frames. This leads to a substantial loss in linguistic meaning  in the generated results. This limitation highlights the importance of the facial enhancement module in sign language video anonymization.

\section{Conclusion and Discussion}



In this paper, we introduce DiffSLVA, a novel approach employing large-scale pre-trained diffusion models for text-guided zero-shot sign language video anonymization in the wild. 

Our approach has the potential to be applied to various use cases. It could enable anonymous peer review for ASL-based academic submissions, thereby ensuring unbiased academic review. Additionally, it could bring neutrality to various multimodal ASL tools, for example, to enable the creation of anonymized definitions in ASL dictionaries. Furthermore, our approach could enhance neutrality in interpreting scenarios in digital communications, such as messaging, enabling maintenance of confidentiality in ASL communications. Furthermore, the implementation of DiffSLVA is likely to increase participation in video-based AI databases, enriching AI research with diverse ASL data.

Our method does currently  have some limitations.  It may encounter challenges, such as cases where the face is occluded by one or both hands or where there is blurring due to rapid movements in sign language videos. We aim to address these issues in our future work. We are also working on further refinements to improve the facial transformation module.

However, overall, DiffSLVA shows substantial promise for anonymization applications in the wild, which could offer invaluable tools for the Deaf and Hard-of-Hearing communities.

\section{Acknowledgments}
We are grateful to the many, many people who have helped with the collection, linguistic annotation, and sharing of the ASL data upon which we have relied for this research.  In particular, we are endebted to the many ASL signers who have contributed to our database; to Gregory Dimitriadis at the Rutgers Laboratory for Computer Science Research, the principal developer of SignStream®, our software for linguistic annotation of video data (\href{https://www.bu.edu/asllrp/SignStream/3/}{https://www.bu.edu/asllrp/SignStream/3/}); to the many who have helped with linguistic annotations (especially Carey Ballard and Indya Oliver); and to Augustine Opoku, for development and maintenance of our Web-based database system for providing access to the linguistically annotated video data 
(\href{https://dai.cs.rutgers.edu/dai/s/dai}{https://dai.cs.rutgers.edu/dai/s/dai}). We would also like to extend our sincere gratitude to Ligong Han for invaluable discussions about this project. This work was supported in part by grants \#2235405, \#2212302, \#2212301, and \#2212303 from the National Science Foundation, although any opinions, findings, and conclusions or recommendations expressed in this material are those of the authors and do not necessarily reflect the views of the National Science Foundation.
\clearpage
\label{sec:discussion}
{
    \small
    \bibliographystyle{ieeenat_fullname}
    \bibliography{main}
}


\end{document}